\documentclass{article}

\usepackage{arxiv}

\usepackage[utf8]{inputenc} 
\usepackage[T1]{fontenc}    
\usepackage{hyperref}       
\usepackage{url}            
\usepackage{booktabs}       
\usepackage{amsfonts}       
\usepackage{nicefrac}       
\usepackage{microtype}      
\usepackage{graphicx}
\usepackage{amsmath}
\usepackage{amssymb}
\usepackage{dirtytalk}
\usepackage[ruled,vlined]{algorithm2e}
\usepackage{authblk}
\usepackage[numbers]{natbib}
\usepackage{doi}

\title{A QuadTree Image Representation for Computational Pathology}

\author{ \href{https://orcid.org/0000-0003-0495-8438}{\includegraphics[scale=0.06]{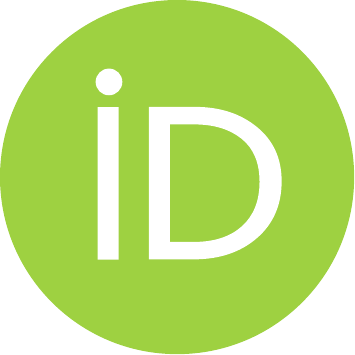}\hspace{1mm}Robert Jewsbury} \hspace{2em}
	\href{https://orcid.org/0000-0001-8830-329X}{\includegraphics[scale=0.06]{orcid.pdf}\hspace{1mm}Abhir Bhalerao} \hspace{2em}
	\href{https://orcid.org/0000-0002-4706-1308}{\includegraphics[scale=0.06]{orcid.pdf}\hspace{1mm}Nasir Rajpoot}
}
\affil[]{TIA Centre, Department of Computer Science, University of Warwick, UK}
\affil[]{\tt\small{\{rob.jewsbury,abhir.bhalerao,n.m.rajpoot\}@warwick.ac.uk}}

\renewcommand{\headeright}{}
\renewcommand{\undertitle}{}
\renewcommand{\shorttitle}{ }

\hypersetup{
pdftitle={A QuadTree Image Representation for Computational Pathology},
pdfsubject={},
pdfauthor={Robert Jewsbury, Abhir Bhalerao, Nasir Rajpoot},
pdfkeywords={},
}

\begin{document}
\maketitle

\begin{abstract}
	The field of computational pathology presents many challenges for computer vision algorithms due to the sheer size of pathology images. Histopathology images are large and need to be split up into image tiles or patches so modern convolutional neural networks (CNNs) can process them. In this work, we present a method to generate an interpretable image representation of computational pathology images using quadtrees and a pipeline to use these representations for highly accurate downstream classification. To the best of our knowledge, this is the first attempt to use quadtrees for pathology image data. We show it is highly accurate, able to achieve as good results as the currently widely adopted tissue mask patch extraction methods all while using over 38\% less data.
\end{abstract}


\section{Introduction}

\label{sec:intro}
Recent advances in digital pathology and the adoption of machine learning methods for computer vision have resulted in many efforts to develop deep learning algorithms for the analysis of information-rich pathology images. The tissue slides analysed by pathologists are very heterogeneous for many reasons including the variety of tissue types, differences in staining protocols and presence of artifacts such as smudges or out of focus regions, to name only a few \cite{QualityCS}. Furthermore, the images are scanned at high resolution, frequently containing hundreds of millions of pixels. Whole Slide Images (WSIs) present a unique challenge in that most state-of-the-art computer vision systems rely on convolutional neural networks (CNNs) \cite{lecun_deep_2015} and are unable to process images larger than a few hundred pixels square in size at once. As such, existing supervised learning methods to analyse large histopathology images typically break them up into small patches losing contextual, global information contained in the larger field of view (FoV) of the entire image \cite{Graham2020DenseSF, Ho2021DeepMN}. Another major limitation of these patch-based methods is that they feed every tissue patch into a CNN, whether it is useful or not.

\subsection{Representation learning in computational pathology}
There has been comparatively little research into novel ways of representing pixel data in pathology images. The vast majority of research has focused on using the patch-based paradigm for a huge variety of tasks from classification through to tumour and tissue segmentation and cell counting regression problems \cite{Litjens2017ASO}. Tellez \textit{et al.} \cite{Tellez2021NeuralIC} propose a two-step method to use CNNs for analysis of gigapixel images using only the image level label termed Neural Image Compression (NIC). They split the input image into adjacent, non-overlapping patches and train an unsupervised compression network to encode semantic information from each patch into a lower-dimensional feature vector. They arrange the extracted feature vectors with the same spatial correspondence as in the original WSI and train a CNN classifier on the extracted feature representation to predict the image level label. With this they are able to achieve an image-level performance only 7\% different compared to a fully supervised baseline. However, the process of creating a representation of the WSI still breaks the image up into patches and assumes the patch level encoder network is able to retain the global information between patches in the generated feature representation. Furthermore, as it is a deep feature representation, it is not easily interpretable and there is no way of verifying whether the extracted feature vectors retain the global information they share with other feature vectors.

\begin{figure*}[t]
\begin{center}
   \includegraphics[width=\linewidth]{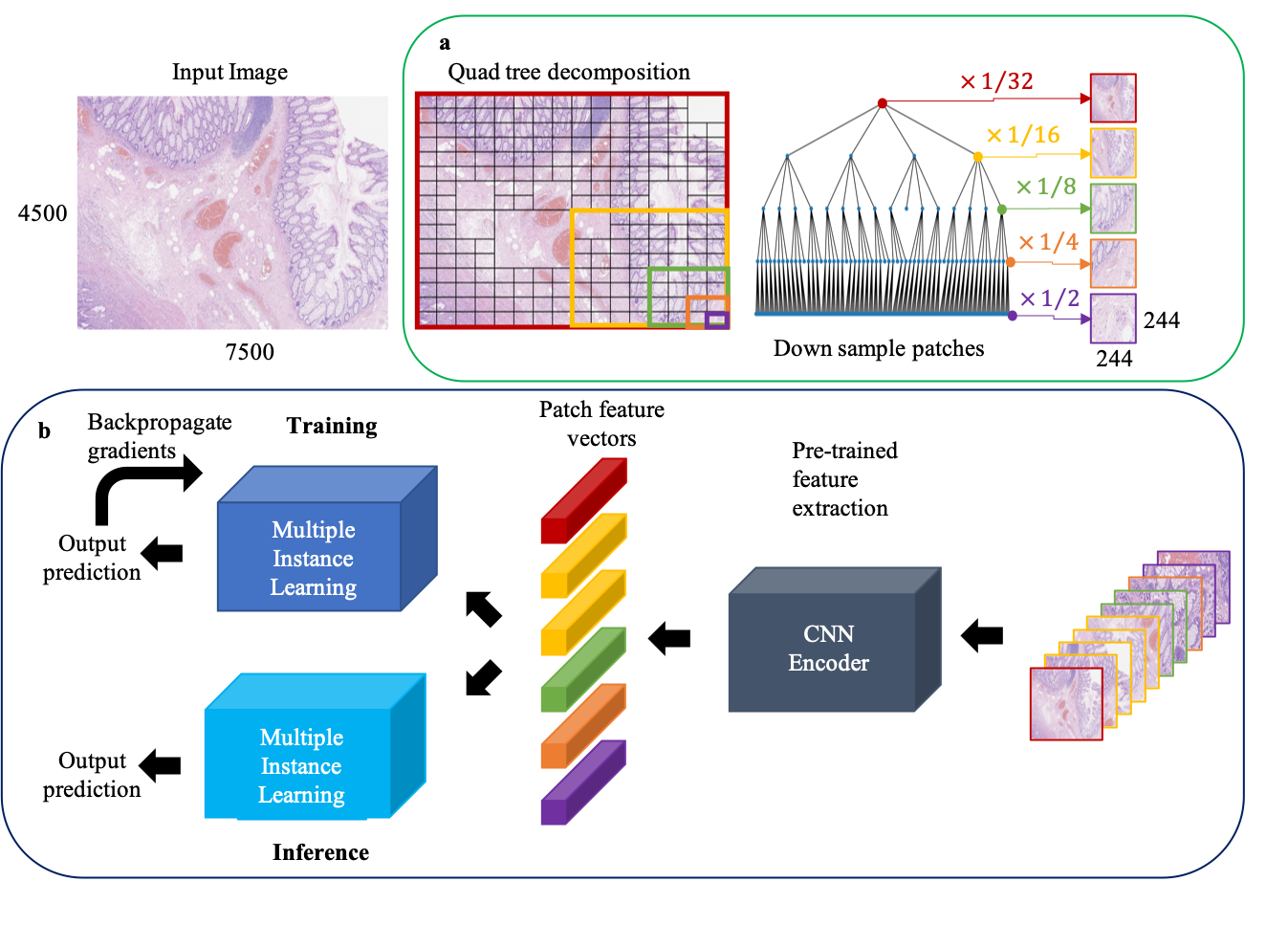}
\end{center}
   \caption{\textbf{Overview of our framework.} \textbf{a} A quadtree is constructed from a given input image, image regions are extracted from each node of the quadtree and down sampled to be of size $244\times 244$ pixels and stored as a bag of patches. \textbf{b} Patches are passed through a feature network. We used a ResNet18 \cite{resnet} pre-trained on ImageNet \cite{imagenet} for all our experiments to encode the feature vectors. These feature vectors are then passed through a MIL framework to generate the final output prediction.}
\label{fig:pipeline}
\end{figure*}

Instead of using a neural network to encode adjacent regions of a histopathology image, we propose to use quadtrees built in a top down fashion from the entire image \cite{quadtrees}. Quadtrees are tree data structures, historically used in image compression \cite{originalQuadtreeImages}. They are constructed by recursively partitioning a two-dimensional space into four sub-regions of equal size and storing the information of each region in a node within a tree. Each node contains information about its corresponding region and has either exactly four children or none in the case of leaf nodes. As such, the tree data structure can be used to represent the information contained in the 2D space in a more compressed, data efficient way. By traversing the tree, the information from the original space can be extracted. The depth of the quadtree can depend on the size of the input image and the distribution of information within the space that is being decomposed. If there is sufficient relevant information within a given region, the algorithm will continue adding nodes to the tree and extending its depth until this no longer holds. The maximum depth of the tree can be limited by truncating the growth at a user defined depth.

Quadtrees have not been used extensively within machine learning. The three-dimensional analogue, octrees, have been explored for tasks such as shape and scene completion with voxel arrays \cite{Wang2017OCNN, Wang2018AdaptiveOA}. Wang \textit{et al.} \cite{Wang2017OCNN} use an Octree-based Convolutional Neural Network (O-CNN) for 3D shape analysis. By building an octree representation of 3D shapes and restricting the computation to the leaf nodes they are able to more efficiently store the information and achieve comparable performance with existing methods while using less memory. Jayaraman \textit{et al.} \cite{QuadTree_CNN} applied this idea with Quadtree Convolutional Neural Network (QCNN) to sparse two-dimensional handwriting data sets resulting in more efficient memory usage and computation time compared to a standard CNN.  

However, Wang \textit{et al.}'s \cite{Wang2017OCNN, Wang2018AdaptiveOA} and Jayaraman \textit{et al.}'s \cite{QuadTree_CNN} work  solely uses binary data. Wang \textit{et al}'s O-CNN works with positional data while Jayaraman \textit{et al}'s QCNN uses greyscale image data. Histopathology data is 2D RGB colour data. In this work, we propose to use the quadtree itself as an image representation while taking advantage of the compression benefits of quadtrees by applying them to RGB data. To our knowledge this is the first time quadtrees have been used for this purpose.

\subsection{Quadtree image representation}

We propose a new image representation built with computational pathology in mind using a type of tree data structure called a quadtree. Unlike existing digital pathology representation methods our framework uses an image representation and is not necessarily a feature representation. We use this quadtree representation to extract patches at varied resolutions in different regions of the images. This leads to greater performance in downstream tasks compared to existing patch extraction techniques all while using less data due to the quadtree construction. We show that our method is able to identify significant regions within an image relevant to clinical diagnosis and generate an interpretable, tree structure representation.

\subsection{Contributions}
Our contributions can be summarised as follows:
\begin{enumerate}
    \item We propose an image representation for computational pathology images and a pipeline able to predict image level labels using a single GPU.
    \item We compare several methods for constructing the quadtree representation using different colour spaces and information quantification functions.
    \item We evaluate our pipeline on a histopathology colorectal adenocarcinoma (CRA) data set used to train the system to classify an image as cancerous or non-cancerous.
    \item We generate attention heatmaps to discover which regions of the image the model found significant in predicting the image level label.
\end{enumerate}

The remainder of the paper is organised as follows: Section \ref{sec:method} explains our method in depth; Section \ref{sec:results} details our experimental results; our discussion and conclusions from our results are stated in Sections \ref{sec:discusion} and \ref{sec:conclusion} respectively.



\section{The Quadtree Framework}
\label{sec:method}
In this section, we present our method for constructing quadtrees from histopathology images and how we use them for downstream analysis. Our pipeline method consists of two main stages.

For a given image, a quadtree is constructed and patches extracted from the tree's nodes. These patches are then treated as a bag of instances and used in a multiple instance learning (MIL) paradigm to generate the image level prediction \cite{Dietterich1997SolvingTM}.

\subsection{Building quadtrees}
\label{sec:building_quad_trees}
A quadtree represents a partition of a two-dimensional space obtained by recursively decomposing the region into four equal quadrants and sub-quadrants where each node in the tree contains information corresponding to a given specific partition of the original space \cite{quadtrees}. In our case the two-dimensional space is that of an RGB image. The quadtree algorithm works on the intuition that if an image or a sub-region within an image, represented by a node in the quadtree, contains sufficient \say{interesting} information it should be divided further into four equally sized sub regions. If this occurs, the tree is expanded by adding four child nodes to the original node being evaluated. However, if a region does not contain sufficient interesting information, then no child nodes are added and the node is made a leaf node. When all regions have been decomposed down to leaf nodes, \textit{i.e.} they do not need to be split any further, then the quadtree is complete. An example of the quadtree representation is shown in Figure \ref{fig:pipeline}a including an example of a decomposed image and its quadtree representation.

\begin{algorithm}
\SetAlgoLined
\SetKwInOut{Input}{Input}
\SetKwInOut{Output}{Output}
\Input{Image: $i\in \mathbb{R}^{m\times n \times 3}$, threshold: $t$, tree depth: $d$, criterion: $c$, quadtree: $Q=$ \textbf{null}}
\Output{quadtree $Q$}

add node $n$ to $Q$ at depth $d$

\If{$n$ has a parent}{
connect parent to $n$
}
\uIf{c($i$) $\leq t$ {\bf or} {\text depth of } $Q = d$}{
    return $Q$
}
\Else{
split $i$ into 4 subquadrants of equal size [$i_1, i_2, i_3, i_4$]

\For{$j$ in [$i_1, i_2, i_3, i_4$\text{]}}{
    QuadTree($j$)
    }
}
 \caption{Quadtree construction algorithm QuadTree}
 \label{alg:quad tree algorithm}
\end{algorithm}

Mathematically, for use with RGB images we represent this idea with the following components:
\begin{enumerate}
    \item A function to quantify the information within a given region which we refer to as the \textit{criterion};
    \item A threshold value which determines the amount of information a region needs to contain to justify splitting the region further.
\end{enumerate}
These components allow us to compute whether the splitting process should occur in a region given the criterion and splitting threshold. 

For example, let us say we have a given region $i\in \mathbb{R}^{m\times n\times 3}$, a splitting threshold $t$ and a criterion function $c$. We can compute $c(i)$ and then compare it with the threshold $t$. If $c(i) > t$ then this means the given region has enough information to justify splitting the region further and expanding the overall quadtree. Conversely, if $c(i) \leq t$ then the region does not have sufficient information to warrant expanding the quadtree and the given node of the tree becomes a leaf node.

\subsubsection{Splitting criteria}
\label{sec:split_method}
We explored several different splitting criterion functions. We used the entropy \cite{Shannon1948AMT} and mean pixel value of an image region as a measure of the amount of significant information in a region. Furthermore, we used these criteria with the images in the native RGB space and two additional colour spaces. All images we used were stained with Haematoxylin and Eosin (H\&E). The haematoxylin dye stains basophilic tissue structures blue, in particular nuclei are heavily stained while cytoplasmic tissue regions are lightly stained. We hypothesise that by focusing on the bluer regions of the image, this will guide the quadtree splitting method to focus on the nuclei within the images which are usually the main feature of interest for most downstream tasks. As such we explored converting the images into blue ratio space \cite{blueRatio} to highlight the areas dense in nuclei. We also used colour deconvolution \cite{colourDeconvolution} to separate out the different stains as colour channels and used the Haematoxylin channel as a proxy for the number of cells.

\subsubsection{Determining the splitting threshold}
\label{sec:split_threshold}
The splitting threshold $t$ used in the quadtree algorithm is a hyperparameter which we determined from the data set. For a given criterion function and colour space, such as entropy in RGB space, we computed the value of this criterion for each image in the data set. We then calculated the mean $(\mu)$ and standard deviation $(\sigma)$ of this distribution and used these values to determine the splitting threshold. We performed an ablation study for this threshold and how it affects both qualitative and downstream performance, as described in Section \ref{sec:res_qtDecomp}.

\subsubsection{Learning from quadtrees}
\label{sec:quad_trees_ml}
In each node of the quadtree we stored a down-sampled copy of the original image region which corresponds to that node's position within the quadtree decomposition. We used image interpolation to down-sample the image regions to all be of size $244\times 244$ for later processing by a pre-trained CNN for feature extraction. This allows us to store patches at different scales from the original image. This incorporates the wider FoV present in the patches corresponding to shallow nodes in the tree while the deeper nodes contain the fine grained detail at high magnification.

\subsection{MIL framework}
\label{sec:mil}
Having generated a collection of down-sampled image regions from the original image using our quadtree method we treated this collection as a \say{bag of instances} or patches. This lends itself to the Multiple Instance Learning (MIL) paradigm, a weakly supervised learning approach, where the bag of instances has a single label but instance level labels are not available \cite{Dietterich1997SolvingTM}. The label of the bag is positive $(Y=1)$ if at least one instance within the bag is positive and conversely the overall bag label is negative $(Y=0)$ if no instance within the bag is positive.

Formally, for a given image $X\in\mathbb{R}^{m\times n \times 3}$ with associated label $Y\in\{0,1\}$ we create a bag $B$ of instances:
\begin{equation}
    B = \{(x_1, y_1), (x_2, y_2), ... , (x_k, y_k)\},
\end{equation}
where $x\in\mathbb{R}^{244\times 244 \times 3}, y\in\{0,1\}$. The instance level labels $y_n$ are hidden to the learner. To classify this bag of images, we can use the following function
\begin{equation}
    \Theta(X) = g(\alpha(f(x_1), ..., f(x_k))),
\end{equation}
where $f$ is a transformation of instances $x_k$ to a lower-dimensional embedding, $\alpha$ is a permutation-invariant aggregation function and $g$ is a transformation to generate the overall class probabilities for the bag.

\begin{figure*}[t]
\begin{center}
   \includegraphics[width=0.8\linewidth]{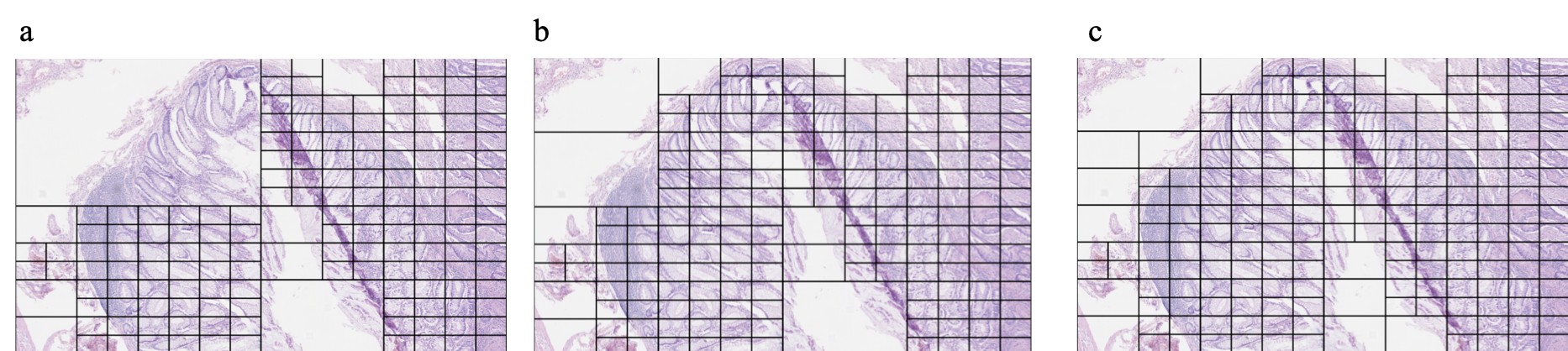}
\end{center}
   \caption{\textbf{Example of quadtree decompositions in different colour spaces}. \textbf{a} Image decomposed using entropy in RGB space. \textbf{b} Image decomposed using mean pixel value in blue ratio space. \textbf{c} Image decomposed using mean pixel value of haematoxylin channel. The splitting threshold was set to the mean ($\mu$) minus one standard deviation ($\sigma$) for each case.}
\label{fig:decompComparison}
\end{figure*}

There are two main approaches to MIL:
\begin{enumerate}
    \item \textit{Instance-based}: Here $f$ classifies each instance individually, assigning a class label for each instance and then $\alpha$ combines these predictions together to generate an overall prediction for the bag using operations such as mean, maximum etc \cite{MILSurvey}. Finally, $g$ is the identity function in this case.
    \item \textit{Embedding-based}: Instead of classifying each instance individually, here $f$ maps each instance to a lower-dimensional embedding, $\alpha$ then obtains a bag representation independent of the number of instances in the bag and $g$ classifies these bag representations to obtain the overall prediction \cite{MILPoolEffects}.
\end{enumerate}
We initially tested the instance-based method but found the classifier was difficult to train to a sufficient accuracy causing poor performance. The embedding-based approach has less bias than the instance-based one and was found to perform better, hence was used for our experiments. For a bag of image patches we used a ResNet-18 pre-trained on ImageNet to encode each image patch as a feature vector turning the bag of image patches into a bag of feature vectors.

\subsubsection{MIL methods}
We used two different MIL methods based on an attention mechanism \cite{attention}. First, we used Attention-based MIL (AMIL) \cite{attentionMIL}. Attention-based MIL modifies the embedding-based MIL approach by changing the MIL pooling operation, $\alpha$, where for a bag $B = \{h_1, ..., h_k\}$ with $K$ embeddings the overall prediction is obtained by
\begin{equation} \label{eq:attention_pred}
    \boldsymbol{z}=\sum_{k=1}^K a_k \boldsymbol{h_k},
\end{equation}
where
\begin{equation} \label{eq:attention}
    a_k=\frac{\exp\{\boldsymbol{w}^T\tanh{\boldsymbol{Vh_k}^T}\}}{\sum_{j=1}^K \exp\{\boldsymbol{w}^T\tanh{\boldsymbol{Vh}_j^T}\}},
\end{equation}
$\boldsymbol{w}\in\mathbb{R}^{L\times1}$ and $\boldsymbol{V}\in\mathbb{R}^{L\times M}$ are parameters. For full details we refer the reader to Ilse \textit{et al.}'s original paper \cite{attentionMIL}.

We also explored a recent expansion of Ilse \textit{et al.}'s Attention-based MIL. Lu \textit{et al.}'s Clustering-constrained Attention Multiple Instance Learning (CLAM) was developed with computational pathology in mind \cite{clam}. In CLAM, the attention network predicts a set of attention scores for each class in the classification problem. Lu \textit{et al.} use the same attention backbone in the first two layers of the network as Ilse \textit{et al.} but then split the network into $n$ parallel attention branches in an $n$-class classification problem along with an additional instance-level clustering layer for each class to obtain the overall prediction. Again, we refer the reader to Lu \textit{et al.}'s original paper for full details of the method \cite{clam}.

\section{Experimental Results}
\label{sec:results}
To evaluate our proposed method we used the CRAG data set from Awan \textit{et al.} \cite{crag}. This contains 139 non-overlapping images extracted from 38 digitised WSIs of colorectal adenocarcinoma (CRA) at 20$\times$ magnification. The images varied slightly but on average were around $4500\times7500$ pixels in size. Of the 139 images, 71 were classified as normal tissue, 33 as low grade and 35 as high grade. We merged the low grade and high grade classes into a cancerous class to create a binary classification problem suitable for the MIL paradigm. In total, we had 71 images in the non-cancerous or negative class and 68 cancerous or positive images creating a fairly balanced classification problem. Awan \textit{et al.} also provided the training/validation split they used for 3-fold cross validation so a fair comparison between our results and theirs could be drawn.

Given the size of the original images, we set the maximum depth $d$ of our quadtrees to be $4$. This was done because if we allowed the trees to create nodes at a deeper level then the original image regions each node represents would be smaller in size than the $244 \times 244$ patch size that are fed into pre-trained networks. Upsampling would have been required to fix this and it would have been inconsistent with the process used on the nodes at every other level within the tree. We kept the trees at depth $4$ for all of our experiments.

During training, we normalised each patch using the mean and standard deviation of ImageNet, augmented each extracted patch using random horizontal and vertical flips with probability 0.5 as well as small, random adjustments to the brightness, contrast, saturation and hue of the input images. We used the Adam optimiser \cite{Adam} with a loss rate of $5\mathrm{e}{-4}$, betas of $0.99$ and $0.999$ and weight decay of $1\mathrm{e}{-4}$. All experiments were performed on a single Nvidia Quadro RTX 5000 GPU.

\subsection{Quadtree decompositions}
\label{sec:res_qtDecomp}
To evaluate the different proposed splitting methods for the quadtrees described in Section \ref{sec:split_method} we performed a qualitative analysis of all 139 images' quadtree decompositions under each method. For each method, we tested a variety of thresholds but kept them consistent in our comparisons. For example, we would only compare images that had been split where the threshold was set to be the same e.g. mean ($\mu$) minus 1 standard deviation ($\sigma$) for each given colour space and method with images split at the same threshold, not with decompositions where the threshold had been set to $\mu - 1.25\sigma$ or any other level.

\begin{table}[t]
\centering
\begin{tabular}{|l|c|c|}
\hline
Method & Accuracy & AUROC\\
\hline\hline
All patches AMIL & $87.83\pm 7.97$ & $0.93\pm0.06$\\
Blue ratio AMIL & $77.01\pm 6.77$ & $0.94\pm0.06$\\
RGB AMIL & $90.72\pm 8.82$ & $0.96\pm 0.04$\\
Haematoxylin AMIL & $94.25\pm 3.31$ & $0.98\pm 0.02$\\\hline
All patches CLAM & $88.55\pm8.60$ &$0.94\pm 0.06$\\
Blue ratio CLAM & $80.54\pm5.88$ & $0.93\pm 0.04$\\
RGB CLAM & $92.06\pm 8.25$ & $0.98\pm0.02$\\
Haematoxylin CLAM & \textbf{97.13 $\pm$ 1.21} & \textbf{0.98 $\pm$ 0.02}\\\hline
\end{tabular}
\caption{Comparison of Attention MIL (AMIL) and CLAM models trained on extracted patches from the 3 different colour spaces at a threshold of $\mu -\sigma$ and with all possible non-overlapping patches from the image. Results reported are \textit{mean} $\pm$ \textit{standard deviation}, the variation in performance occurs due to differences in cross validation performance}
\label{table:splitting_method_ablation}
\end{table}


We found that the mean pixel value in the Haematoxylin channel gave the best qualitative decomposition of the input images. Figure \ref{fig:decompComparison} shows an example of this. The method consistently split the image into finer grained detail in the presence of tissue while ignoring the background or less informative regions such as empty space or large areas of adipose tissue. Furthermore, it would aggressively split regions in the presence of cancerous tissue while with non cancerous tissue the method would sometimes only split down to a depth of 2 or 3 in more homogeneous areas such as large regions of connective tissue. This was not found to be the case in large regions of cancerous tissue where the method would consistently split the image down to the maximum allowed depth. Additionally, the threshold of $\mu - \sigma$ was empirically found to give the best qualitative splits by consistently following and splitting the tissue regions within the images.

We trained the Attention MIL \cite{attentionMIL} and CLAM \cite{clam} models for 20 epochs using the patches extracted with each different method. As a baseline control comparison, we also divided every image into non-overlapping patches of size $244 \times 244$ denoted as the \say{all patches} baseline. The results in Table \ref{table:splitting_method_ablation} show that our qualitative findings that the Haematoxylin channel method produced the best decompositions of the images also corresponded to the best downstream model performance with both Attention MIL and CLAM when compared to the other methods with all other variables kept constant. Furthermore, two of the three colour spaces, RGB and Haematoxylin channel, outperformed our baseline of using every possible patch in an image as a bag of instances. The third colour space, blue ratio space, performed relatively poorly as it almost always predicted the positive class giving a very high sensitivity but leading to overall worse performance than the other methods. Additionally, it was found that in each respective colour space and with the all patches baseline the CLAM model outperformed the Attention MIL model in terms of average performance with a smaller standard deviation for all cases except the all patches baseline.

\begin{figure}[t]
\begin{center}
   \includegraphics[width=0.7\linewidth]{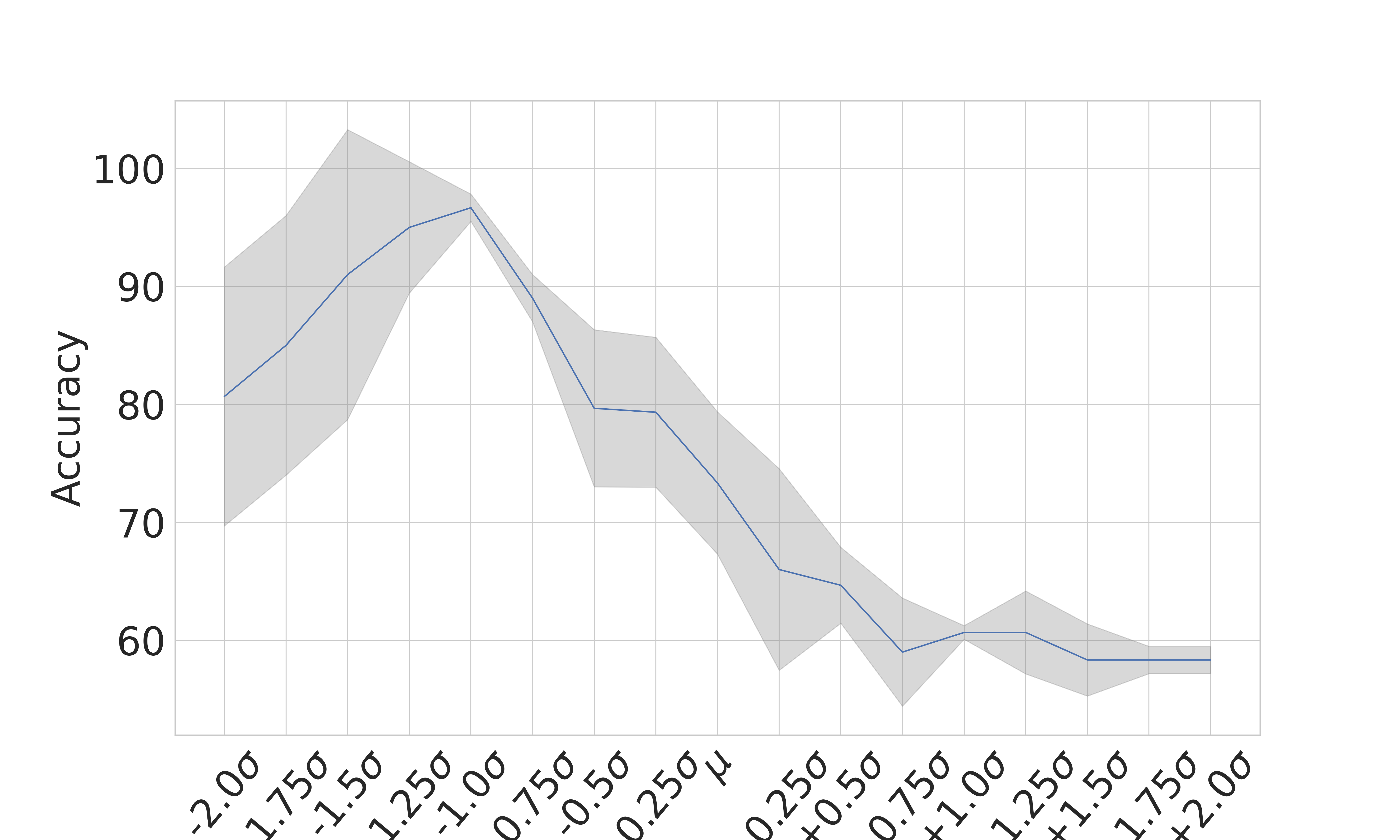}
\end{center}
   \caption{Accuracy of a CLAM model trained on data sets extracted at different thresholds ranging from $\mu - 2\sigma$ to $\mu + 2\sigma$ in the Haematoxylin channel using the mean pixel value. Grey area indicates $\pm$ one standard deviation in the model's performance across 3 fold cross validation.}
\label{fig:ablationAccuracies}
\end{figure}

\subsection{Ablation study}
We performed an ablation study of the splitting thresholds used with the best performing approach, using the mean pixel value in the Haematoxylin channel. We varied the threshold from $\mu -2\sigma$ to $\mu +2\sigma$ in steps of $0.25\sigma$ and trained a CLAM model with 3 fold cross validation at each of the 17 thresholds, all other hyperparameters were kept constant. We found the model had the greatest performance using the best qualitative threshold of $\mu -\sigma$ as it had the highest average performance and the smallest standard deviation across cross validation folds as shown in Figure \ref{fig:ablationAccuracies}.

\begin{table*}[b]
\begin{center}
\begin{tabular}{|l|c|c|c|c|}
\hline
Method & Accuracy & AUROC & Average Training Time (s) & \% of pixel data used in training\\
\hline\hline
BAM 1 \cite{crag} & $95.70\pm 2.10$ & $-$ & $-$ & 100\\
BAM 2 \cite{crag} & $97.12 \pm 1.27$ & $-$ & $-$ & 100\\
\hline\hline
All patches & $88.55\pm8.60$ & $0.94\pm 0.06$ & $2110$ & 100\\
Segmented patches & $97.15\pm 3.24$ & \textbf{1.00 $\pm$ 0.01} & $2097$ & $69$\\
Leaf nodes & $95.65\pm 2.11$ & $0.99 \pm 0.02$ & \textbf{1041} & \textbf{32}\\
All nodes & \textbf{97.13 $\pm$ 1.21} & $0.98\pm 0.02$ & $1368$ & $43$\\
\hline
\end{tabular}
\end{center}
\caption{Performance of CLAM model trained on 4 different sets of extracted patches compared to state-of-the-art from original CRAG paper on two class problem. Haematoxylin channel patches were extracted using the threshold $\mu -\sigma$. Reported metric values are \textit{mean $\pm$ standard deviation} on the validation set, the variation is due to the 3-fold cross validation. Training times are averaged across the 3 cross validation folds for 20 epochs with all hyperparameters kept constant.}
\label{table:clam_ablation}
\end{table*}

\subsection{Comparisons with other methods}
We performed an additional comparison using the standard practice for extracting patches in tissue regions in computational pathology. To select only regions with tissue we thresholded an image's intensity to separate the tissue from the background and created a tissue mask with Otsu's method \cite{Otsu}. We obtained a set of locations within the tissue area to extract patches from with the super-pixel algorithm such that the locations covered the entire tissue region \cite{slic}. To explore the significance of the non-leaf nodes with the quadtree structure in downstream tasks we also compared performance using just the leaf nodes patches extracted by the Haematoxylin channel method.

Table \ref{table:clam_ablation} shows that the patches extracted by our Haematoxylin channel quadtree method yield as good performance as the segmented tissue patches with a smaller standard deviation between cross validation folds. In total the segmentation method extracted 52,029 patches, 374 per image, while the Haematoxylin channel quadtree method extracted 32,099, 231 per image, a 38.31\% reduction in data which still yielded as good if not better final model performance with all other training hyperparameters the same.

\begin{figure}
\begin{center}
  \includegraphics[width=0.7\linewidth]{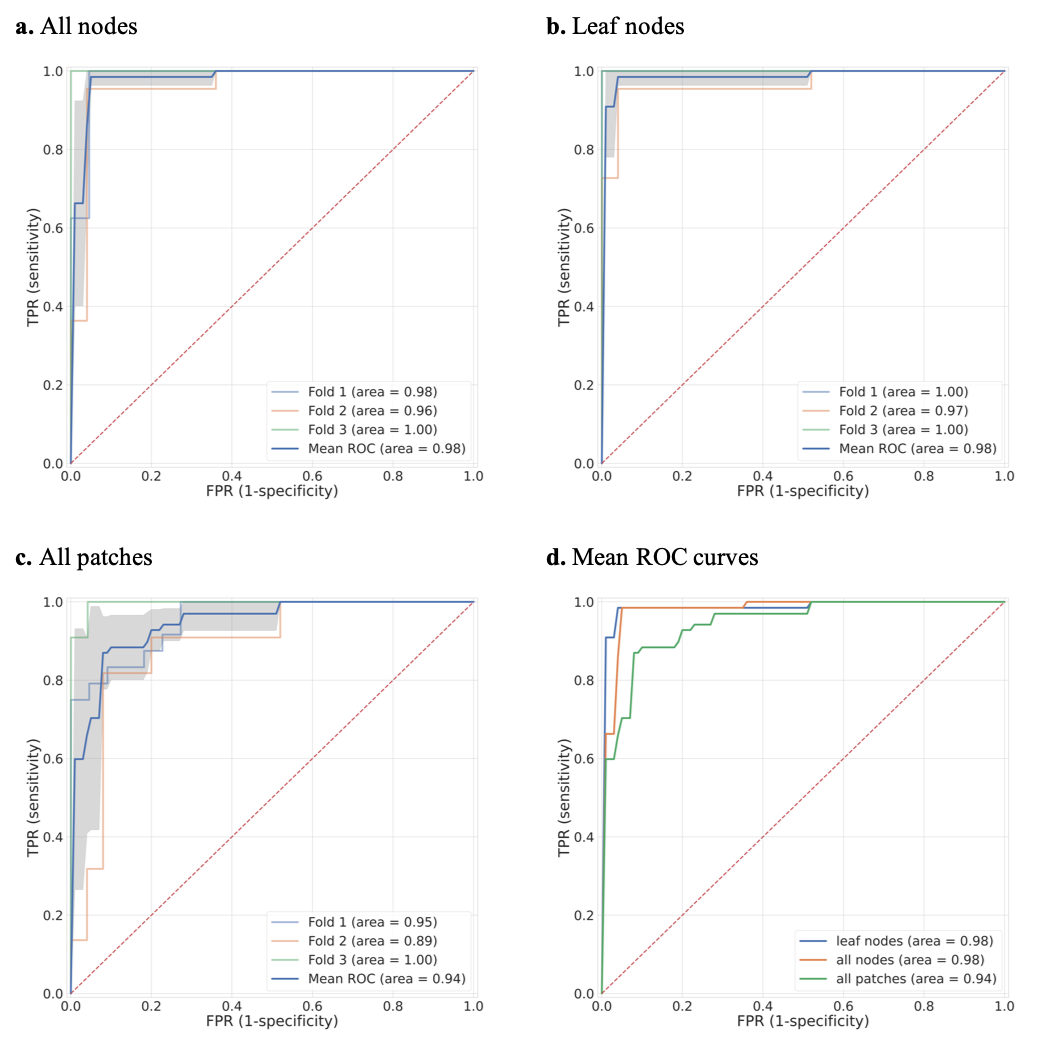}
\end{center}
  \caption{\textbf{a, b, c} AUROC curve plots of cross validation average AUROC for CLAM models trained on all quadtree nodes, the leaf nodes from the quadtree and all patches respectively. The quadtree used was the best performing haematoxylin channel method extracted at a threshold of $\mu - \sigma$. Grey shaded area indicates $\pm$ one standard deviation across the three folds. \textbf{d} Average AUROC curve for all three methods}
\label{fig:rocs}
\end{figure}

\section{Discussion}
\label{sec:discusion}

\begin{figure*}[t]
\begin{center}
   \includegraphics[width=0.9\linewidth]{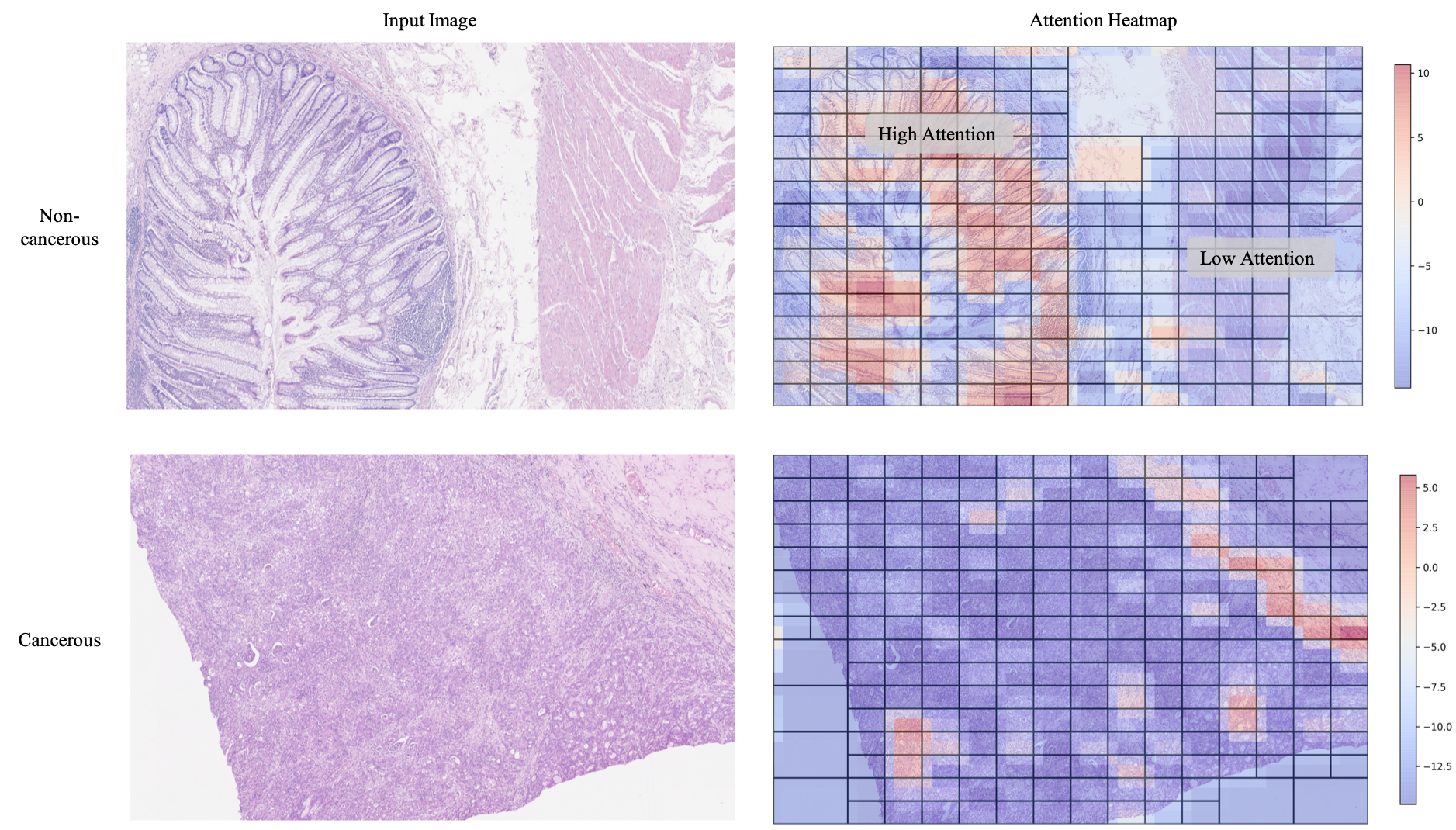}
\end{center}
   \caption{\textbf{Attention heatmap visualisations} For a non-cancerous (top) and a cancerous (bottom) image respectively the attention heatmap from the CLAM model trained using the Haematoxylin channel extracted patches at $\mu -\sigma$ overlaid on the quadtree decomposition of the original image (right). We smoothed the attention heatmaps for visibility by using 50\% overlap between the patches and averaged the attention weights. The most highly attended regions are denoted in red while the less attended regions are denoted in blue. Regions have been re-scaled back to their original shape and size for display purposes.}
\label{fig:attentionHeatmap}
\end{figure*}

Figure \ref{fig:ablationAccuracies} shows that our qualitative finding of setting the splitting threshold at $\mu - \sigma$ yielded the best results also held true when compared to a wide range of possible thresholds in downstream performance. When the splitting threshold was set too high e.g. $t > (\mu + \sigma)$ we found that frequently the non-cancerous class images would not be split at all resulting in a quadtree with one root node only and a corresponding bag containing a single instance. This occurred because the non-cancerous images had more empty space on average than the cancerous images and contained less regions heavily stained with Haematoxylin. This in turn led to poor performance as the model was operating on a single instance which was heavily downsampled and did not have enough data to learn from and perform well. As this threshold was decreased the images were split more effectively by the algorithm resulting in improved performance. The performance peaked at the qualitatively best performing threshold of $\mu - \sigma$ with the smallest standard deviation between cross validation folds. The performance then decreased beyond this point with a much higher variation across cross validation folds. Inspecting the individual fold performance showed this was driven by one of the three folds performing much poorer than the others; however, this was not always the same fold in each case so this cannot have been occurring due to a difference in the data distribution between folds. 

We found that when $t < \mu - \sigma$ the algorithm started to split regions that contained a low percentage of tissue where it was not before. This resulted in one patch containing tissue being generated but potentially two or three containing majority or all blank space. We believe this is why performance becomes more varied, more instances are being generated that are not conducive to making an accurate prediction. The threshold of $\mu - \sigma$ appears to be a good middle ground between decomposing the tissue regions accurately and not over decomposing the images. While it is logical that such a middle ground should exist theoretically, we do not yet know why this is found at $\mu - \sigma$. We plan to explore this approach with other data sets in future to see if this finding holds true in different data distributions.

We found that our quadtree image representation is able to achieve better performance than the existing state of the art method for patch extraction in computational pathology. The CLAM model trained with the patches extracted from our quadtree method was able to achieve as good average cross validation performance with a smaller standard deviation compared to a model trained using patches extracted using a tissue segmentation mask while using 38.31\% less patches. 

When we excluded the non-leaf nodes' patches, the final model performance did degrade slightly but only by ~1\% compared to when all patches from the quadtree were included. We hypothesise this indicates that the higher level nodes in the tree (at lower magnifications) only contribute marginally to the final prediction. To verify this we overlaid the attention weights from the trained CLAM model in a heatmap over the original images. If the higher level non-leaf nodes were weighted highly we would expect to see large regions of high attention overlaid over regions that have been split further. However, the example in Figure \ref{fig:attentionHeatmap} does not show this. It is representative of many other instances in our results where the leaf nodes are clearly the most highly weighted.

The attention heatmap in Figure \ref{fig:attentionHeatmap} also shows that the CLAM model trained using our quadtree strategy is able to assign higher attention values to clinically relevant regions. In the non-cancerous example, almost all of the highly weighted patches are located in a region of healthy glands and tissue, highlighted in red and orange, while the stromal region has been coloured a dark blue indicating very low attention has been assigned in this region by the model. This means the model has correctly identified that the presence of healthy glands indicates a non-cancerous region.

\section{Conclusion}
\label{sec:conclusion}
We have shown that quadtrees can be used in computational pathology as an efficient image representation which can be used for fast and highly accurate downstream performance. Our quadtree method is able to decompose histopathology images by identifying regions significant to clinical diagnosis while ignoring less significant regions such as empty space or connective tissue. We show that this image representation is able to be leveraged in a MIL setting to achieve better performance using 38\% less data than the currently widely adopted thresholding based tissue mask approach used in the field while also providing an interpretable visualisation of which regions within the image are important for the algorithm in generating its prediction.

In future, we plan to explore this approach with other computational pathology data sets to further verify our findings and explore how the framework performs for other tasks and tissue types as well as how sensitive it is to other staining methods and visual artifacts. If our results here hold true, the method should be very helpful in WSIs to reduce the significant data requirement currently present in processing these large images.

\bibliographystyle{IEEEtran}
\bibliography{QuadTree}

\begin{thebibliography}{10}
\providecommand{\url}[1]{#1}
\csname url@samestyle\endcsname
\providecommand{\newblock}{\relax}
\providecommand{\bibinfo}[2]{#2}
\providecommand{\BIBentrySTDinterwordspacing}{\spaceskip=0pt\relax}
\providecommand{\BIBentryALTinterwordstretchfactor}{4}
\providecommand{\BIBentryALTinterwordspacing}{\spaceskip=\fontdimen2\font plus
\BIBentryALTinterwordstretchfactor\fontdimen3\font minus
  \fontdimen4\font\relax}
\providecommand{\BIBforeignlanguage}[2]{{%
\expandafter\ifx\csname l@#1\endcsname\relax
\typeout{** WARNING: IEEEtran.bst: No hyphenation pattern has been}%
\typeout{** loaded for the language `#1'. Using the pattern for}%
\typeout{** the default language instead.}%
\else
\language=\csname l@#1\endcsname
\fi
#2}}
\providecommand{\BIBdecl}{\relax}
\BIBdecl

\bibitem{QualityCS}
B.~Sch{\"o}mig-Markiefka, A.~Pryalukhin, W.~Hulla, A.~Bychkov, J.~Fukuoka,
  A.~Madabhushi, V.~Achter, L.~Nieroda, R.~B{\"u}ttner, A.~Quaas, and
  Y.~Tolkach, ``Quality control stress test for deep learning-based diagnostic
  model in digital pathology,'' \emph{Modern Pathology}, pp. 1 -- 11, 2021.

\bibitem{lecun_deep_2015}
\BIBentryALTinterwordspacing
Y.~LeCun, Y.~Bengio, and G.~Hinton, ``Deep learning,'' \emph{Nature}, vol. 521,
  no. 7553, pp. 436--444, May 2015. [Online]. Available:
  \url{https://doi.org/10.1038/nature14539}
\BIBentrySTDinterwordspacing

\bibitem{Graham2020DenseSF}
S.~Graham, D.~Epstein, and N.~Rajpoot, ``Dense steerable filter cnns for
  exploiting rotational symmetry in histology images,'' \emph{IEEE Transactions
  on Medical Imaging}, vol.~39, pp. 4124--4136, 2020.

\bibitem{Ho2021DeepMN}
D.~J. Ho, D.~V.~K. Yarlagadda, T.~D’Alfonso, M.~Hanna, A.~Grabenstetter,
  P.~Ntiamoah, E.~Brogi, L.~Tan, and T.~J. Fuchs, ``Deep multi-magnification
  networks for multi-class breast cancer image segmentation,''
  \emph{Computerized medical imaging and graphics : the official journal of the
  Computerized Medical Imaging Society}, vol.~88, p. 101866, 2021.

\bibitem{Litjens2017ASO}
G.~Litjens, T.~Kooi, B.~E. Bejnordi, A.~Setio, F.~Ciompi, M.~Ghafoorian,
  J.~V.~D. Laak, B.~Ginneken, and C.~S{\'a}nchez, ``A survey on deep learning
  in medical image analysis,'' \emph{Medical image analysis}, vol.~42, pp.
  60--88, 2017.

\bibitem{Tellez2021NeuralIC}
D.~Tellez, G.~Litjens, J.~A. van~der Laak, and F.~Ciompi, ``Neural image
  compression for gigapixel histopathology image analysis,'' \emph{IEEE
  Transactions on Pattern Analysis and Machine Intelligence}, vol.~43, pp.
  567--578, 2021.

\bibitem{resnet}
K.~He, X.~Zhang, S.~Ren, and J.~Sun, ``Deep residual learning for image
  recognition,'' \emph{2016 IEEE Conference on Computer Vision and Pattern
  Recognition (CVPR)}, pp. 770--778, 2016.

\bibitem{imagenet}
J.~Deng, W.~Dong, R.~Socher, L.-J. Li, K.~Li, and L.~Fei-Fei, ``Imagenet: A
  large-scale hierarchical image database,'' in \emph{2009 IEEE Conference on
  Computer Vision and Pattern Recognition}, 2009, pp. 248--255.

\bibitem{quadtrees}
R.~Finkel and J.~Bentley, ``Quad trees a data structure for retrieval on
  composite keys,'' \emph{Acta Informatica}, vol.~4, pp. 1--9, 1974.

\bibitem{originalQuadtreeImages}
S.~Tanimoto and T.~Pavlidis, ``A hierarchical data structure for picture
  processing,'' \emph{Computer Graphics and Image Processing}, vol.~4, pp.
  104--119, 1975.

\bibitem{Wang2017OCNN}
P.-S. Wang, Y.~Liu, Y.-X. Guo, C.-Y. Sun, and X.~Tong, ``O-cnn,'' \emph{ACM
  Transactions on Graphics (TOG)}, vol.~36, pp. 1 -- 11, 2017.

\bibitem{Wang2018AdaptiveOA}
P.-S. Wang, C.-Y. Sun, Y.~Liu, and X.~Tong, ``Adaptive o-cnn: A patch-based
  deep representation of 3d shapes,'' \emph{arXiv: Computer Vision and Pattern
  Recognition}, 2018.

\bibitem{QuadTree_CNN}
P.~Jayaraman, J.~Mei, J.~Cai, and J.~Zheng, ``Quadtree convolutional neural
  networks,'' in \emph{ECCV}, 2018.

\bibitem{Dietterich1997SolvingTM}
T.~G. Dietterich, R.~Lathrop, and T.~Lozano-Perez, ``Solving the multiple
  instance problem with axis-parallel rectangles,'' \emph{Artif. Intell.},
  vol.~89, pp. 31--71, 1997.

\bibitem{Shannon1948AMT}
C.~Shannon, ``A mathematical theory of communication,'' \emph{Bell Syst. Tech.
  J.}, vol.~27, pp. 379--423, 1948.

\bibitem{blueRatio}
G.~Lu, D.~Wang, X.~Qin, S.~Muller, J.~V. Little, X.~Wang, A.~Y. Chen, G.~Chen,
  and B.~Fei, ``Histopathology feature mining and association with
  hyperspectral imaging for the detection of squamous neoplasia,''
  \emph{Scientific Reports}, vol.~9, 2019.

\bibitem{colourDeconvolution}
A.~Ruifrok and D.~Johnston, ``Quantification of histochemical staining by color
  deconvolution.'' \emph{Analytical and quantitative cytology and histology},
  vol. 23 4, pp. 291--9, 2001.

\bibitem{MILSurvey}
M.~Carbonneau, V.~Cheplygina, E.~Granger, and G.~Gagnon, ``Multiple instance
  learning: A survey of problem characteristics and applications,''
  \emph{Pattern Recognit.}, vol.~77, pp. 329--353, 2018.

\bibitem{MILPoolEffects}
M.~U. Oner, J.~M.~S. Kye-Jet, H.~Lee, and W.-K. Sung, ``Studying the effect of
  mil pooling filters on mil tasks,'' \emph{ArXiv}, vol. abs/2006.01561, 2020.

\bibitem{attention}
D.~Bahdanau, K.~Cho, and Y.~Bengio, ``Neural machine translation by jointly
  learning to align and translate,'' \emph{CoRR}, vol. abs/1409.0473, 2015.

\bibitem{attentionMIL}
M.~Ilse, J.~M. Tomczak, and M.~Welling, ``Attention-based deep multiple
  instance learning,'' in \emph{ICML}, 2018.

\bibitem{clam}
M.~Lu, D.~F.~K. Williamson, T.~Y. Chen, R.~J. Chen, M.~Barbieri, and
  F.~Mahmood, ``Data efficient and weakly supervised computational pathology on
  whole slide images,'' \emph{Nature biomedical engineering}, 2021.

\bibitem{crag}
R.~Awan, K.~Sirinukunwattana, D.~Epstein, S.~D.~R. Jefferyes, U.~Qidwai,
  Z.~Aftab, I.~Mujeeb, D.~Snead, and N.~Rajpoot, ``Glandular morphometrics for
  objective grading of colorectal adenocarcinoma histology images,''
  \emph{Scientific Reports}, vol.~7, 2017.

\bibitem{Adam}
D.~P. Kingma and J.~Ba, ``Adam: A method for stochastic optimization,''
  \emph{CoRR}, vol. abs/1412.6980, 2015.

\bibitem{Otsu}
N.~Otsu, ``A threshold selection method from gray level histograms,''
  \emph{IEEE Transactions on Systems, Man, and Cybernetics}, vol.~9, pp.
  62--66, 1979.

\bibitem{slic}
R.~Achanta, A.~Shaji, K.~Smith, A.~Lucchi, P.~Fua, and S.~S{\"u}sstrunk, ``Slic
  superpixels compared to state-of-the-art superpixel methods,'' \emph{IEEE
  Transactions on Pattern Analysis and Machine Intelligence}, vol.~34, pp.
  2274--2282, 2012.

\end{thebibliography}






\end{document}